\documentclass[10pt,journal,compsoc]{IEEEtran}

\usepackage[compress]{cite}
\usepackage{microtype}
\usepackage{graphicx}
\usepackage{subfigure}
\usepackage{booktabs} 
\usepackage[hidelinks]{hyperref}

\usepackage{amsmath,amssymb,amsthm}
\usepackage{dsfont}
\usepackage{tikz}
\usepackage{psfrag}
\usepackage{algorithm}
\usepackage{algorithmic}

\PassOptionsToPackage{usenames,dvipsnames}{xcolor}

\definecolor{TealBlue}{rgb}{0.184, 0.496, 0.463}

\usetikzlibrary{arrows,positioning,automata,calc,shapes}

\definecolor{myblue}{rgb}{0.0000,0.4470,0.7410}
\definecolor{myred}{rgb}{0.8500,0.3250,0.0980}
\definecolor{myorange}{rgb}{0.9290,0.6940,0.1250}

\newtheorem{assumption}{Assumption}
\newtheorem{theorem}{Theorem}
\newtheorem{definition}{Definition}

\newcommand{\farhad}[1]{{\color{black}#1}}

\DeclareMathOperator*{\argmin}{arg\,min}

\DeclareMathOperator{\diam}{diam}

\makeatletter
\pgfdeclareshape{document}{
\inheritsavedanchors[from=rectangle] 
\inheritanchorborder[from=rectangle]
\inheritanchor[from=rectangle]{center}
\inheritanchor[from=rectangle]{north}
\inheritanchor[from=rectangle]{south}
\inheritanchor[from=rectangle]{west}
\inheritanchor[from=rectangle]{east}
\backgroundpath{
\southwest \pgf@xa=\pgf@x \pgf@ya=\pgf@y
\northeast \pgf@xb=\pgf@x \pgf@yb=\pgf@y
\pgf@xc=\pgf@xb \advance\pgf@xc by-10pt 
\pgf@yc=\pgf@yb \advance\pgf@yc by-10pt
\pgfpathmoveto{\pgfpoint{\pgf@xa}{\pgf@ya}}
\pgfpathlineto{\pgfpoint{\pgf@xa}{\pgf@yb}}
\pgfpathlineto{\pgfpoint{\pgf@xc}{\pgf@yb}}
\pgfpathlineto{\pgfpoint{\pgf@xb}{\pgf@yc}}
\pgfpathlineto{\pgfpoint{\pgf@xb}{\pgf@ya}}
\pgfpathclose
\pgfpathmoveto{\pgfpoint{\pgf@xc}{\pgf@yb}}
\pgfpathlineto{\pgfpoint{\pgf@xc}{\pgf@yc}}
\pgfpathlineto{\pgfpoint{\pgf@xb}{\pgf@yc}}
\pgfpathlineto{\pgfpoint{\pgf@xc}{\pgf@yc}}
}
}
\makeatother

\begin{document}

\title{The Cost of Privacy in Asynchronous Differentially-Private Machine Learning}

\author{Farhad~Farokhi,~\IEEEmembership{Senior Member,~IEEE}, Nan Wu, David Smith, and Mohamed Ali Kaafar

	\IEEEcompsocitemizethanks{
		\IEEEcompsocthanksitem F.~Farokhi is with the Department of Electrical and Electronic Engineering at the University of Melbourne, Australia.\protect\\ E-mail: Farhad.Farokhi@unimelb.edu.au
		\IEEEcompsocthanksitem N.~Wu is with the Macquarie University and CSIRO's Data61, Australia.\protect\\ E-mail: Nan.Wu@data61.csiro.au
		\IEEEcompsocthanksitem D.~Smith is with the CSIRO's Data61 and the Australian National University, Australia.\protect\\ E-mail: David.Smith@data61.csiro.au
		\IEEEcompsocthanksitem M. A.~Kaafar is with the Macquarie University and CSIRO's Data61, Australia.\protect\\ E-mail: Dali.Kaafar@data61.csiro.au
	}
	\thanks{Manuscript received \today.}
}


\IEEEtitleabstractindextext{%
	\begin{abstract}
		We consider training machine learning models using data located on multiple private and geographically-scattered servers with different privacy settings. Due to the distributed nature of the data, communicating with all collaborating private data owners simultaneously may prove challenging or altogether impossible. We consider differentially-private asynchronous algorithms for collaboratively training machine-learning models on multiple private datasets. The asynchronous nature of the algorithms implies that a central learner interacts with the private data owners one-on-one whenever they are available for communication without needing to aggregate query responses to construct gradients of the entire fitness function. Therefore, the algorithm efficiently scales to many data owners. We define the cost of privacy as the difference between the fitness of a privacy-preserving machine-learning model and the fitness of trained machine-learning model in the absence of privacy concerns. We demonstrate that the cost of privacy has an upper bound that is inversely proportional to the combined size of the  training datasets squared and the sum of the privacy budgets squared. We validate the theoretical results with experiments on financial and medical datasets. The experiments illustrate that collaboration among more than 10 data owners with at least 10,000 records with privacy budgets greater than or equal to 1 results in a superior machine-learning model in comparison to a model trained in isolation on only one of the datasets, illustrating the value of collaboration and the cost of the privacy. The number of the collaborating datasets can be lowered if the privacy budget is higher.
	\end{abstract}
	
	\begin{IEEEkeywords}
		Machine learning; Differential privacy; Stochastic gradient algorithm; Asynchronous.
\end{IEEEkeywords}}

\maketitle

\IEEEdisplaynontitleabstractindextext

\IEEEpeerreviewmaketitle

\IEEEraisesectionheading{\section{Introduction}}
\IEEEPARstart{U}{nprecedented} abundance of data has ignited a machine learning (ML) race that aims to boost productivity and spur economic growth globally. However, the data required for training such ML models is often distributed across multiple independent competing entities, e.g., financial or energy data is often scattered across servers for several service providers with competing interests. Regulatory frameworks, such as the GDPR, are increasingly restricting migration of private data across companies or even geographical boundaries for possible merger and training. This might restrict ML techniques from accessing the data in its entirety for training models, which motivates the development of distributed ML techniques with privacy guarantees.

Training data for machine learning can be located on multiple private geographically-scattered servers with different privacy settings. For instance, the training data can be gathered by Internet of Things (IoT) devices or hosted locally on smart devices with privacy settings enforced by users. Another example is cross-sector or -services ML with cross-governance datasets. In these cases, communicating with all private data owners simultaneously when training ML models is unpractical, if not impossible. A learner (i.e., a central agent responsible for training ML models) needs to resort to asynchronous communication with the different data owners. This implies that the learner can communicate with the data owners on a one-on-one basis without needing to wait for all data owners to respond. When using a gradient descent algorithm for training the ML model, the asynchronous communication raises an important challenge: the learner no longer knows the direction for the best model update based on all the training dataset; it can only infer the best update direction for the communicating data owner.

In this paper, we investigate the fitness of asynchronous ML learning algorithms. 
The learner updates the model based on differentially-private (DP)~\cite{dwork2006calibrating,dwork2014algorithmic} gradient of only the part of the fitness that depends on the data possessed by the communicating data owner. To address the challenge of not knowing the direction for the best model update, the learner updates the ML model with small, yet constant, learning rates. The learner also shows inertia in updating its ML model so that it does not change the model significantly because of the gradient of just one data owner. These choices are motivated by that the learner is not overly confident that an update that is good for one data owner is also good for the others. The constant learning rate and the inertia of the learner allow the gradients of all the data owners to get mixed with across time so that the learner follows the direction for the best model update.

Note that, in this paper, we only investigate honest-but-curious threats in which the data owners do not trust the learner or each other for sharing private  datasets while they trust that the learner trains the model correctly based on a specified algorithm. For instance, in a financial sector, a  central bank or a government organisation may be trusted for training ML models from distributed datasets but financial organisations prefer not to share their original data with the bank nor with each other. Likewise, in the health sector, a government organisation may be trusted to play the role of the central learner. For more general settings, incentives must be provided to ensure that the learner follows the training algorithm~\cite{parkes2004distributed,tanaka2015faithful}.

The difference between the fitness function evaluated for privacy-preserving ML model and the fitness function evaluated for trained ML model without privacy concerns, or the degradation caused in the performance of ML models by the presence of differential privacy noise, captures  the cost of privacy. In this paper, we prove that the cost of privacy is inversely proportional to the combined size of the  training datasets squared and the privacy budgets squared. We validate the theoretical results on experiments on financial data. We use linear  regression on a dataset of loan information from the Lending Club, a peer-to-peer lending platform, for setting interest rates of loans based on attributes, such as loan size and credit rating. We also use regression models on a dataset of hospital visits by patients in the U.S for determining the length of stay based on attributes, such as age, gender, and diagnosis. We show that, for collaboration among large numbers of private data owners, i.e., more than 10 data owners with at least 10,000 records, and with relatively large privacy budgets, i.e., privacy budgets greater than 1, the performance of the private ML model can beat the performance of a model that is trained with no collaboration. Therefore, we establish the value of collaboration in ML between multiple private data owners.


\section{Related Work}

\textit{ML with Differential Privacy}: 
Previous work ~\cite{sarwate2013signal,zhang2012functional,chaudhuri2009privacy,zhang2016differential,duchi2013local} studied ML training under the differential privacy framework. These approaches require merging the private datasets for training and rely on obfuscating the generated ML model using DP once the training on the aggregated data is performed. Alternatively, an ML model based on the obfuscated, yet merged data is trained. These studies do not consider the need for privacy preservation prior to merging the data. In addition they \textit{do not consider the asynchronous nature of the communication between the learner and the data owners} by only requiring responses to some queries on the private dataset.

\textit{Distributed/Collaborative Privacy-Preserving ML}:
Distributed privacy-preserving ML proposes the use of DP gradients for training ML models~\cite{zhang2017dynamic,huang2018dp,abadi2016deep,zhang2018privacy,farokhi_ieee_sp_2020,mcmahan2017learning,shokri2015privacy,papernot2018scalable,pate_2017,smith2017interaction}. 
Noisy DP gradients can be used to train ML models with convex and non-convex fitness functions~\cite{farokhi_ieee_sp_2020,shokri2015privacy,mcmahan2017learning}. An important aspect of these studies is that they sometimes use better DP composition methods, such as moment accountant, for reducing the scale of the DP noise~\cite{abadi2016deep}. \textit{These studies  however propose synchronous updates} in which the ML model must be updated according to the contributions of all the data owners simultaneously (rather than a subset of them). This assumption can prohibit the use of the above distributed or collaborative ML training algorithms in the presence of numerous data owners. They also \textit{do not provide a forecast for the performance of privacy-preserving trained models}. \farhad{Note that this paper addresses a similar problem to~\cite{farokhi_ieee_sp_2020} but under an asynchronous model by allowing the learner to communicate with the datasets in a one-on-one basis whenever they are available.} The availability for communication is particularly modelled using Poisson point processes. These processes are often utilized for analysis of asynchronous multi-agent systems and are shown to mimic practical scenarios~\cite{ram2009asynchronous,heidelberger1982queueing,lagunoff1997asynchronous}.

\textit{Asynchronous Distributed Optimization and ML}:  Distributed asynchronous optimization algorithms can be used for training ML models~\cite{srivastava2011distributed,5454103,aybat2015asynchronous,ram2009asynchronous,bedi2018asynchronous,hong2017stochastic}. This is because we can formulate distributed ML training as a distributed optimization problem with private datasets represented as parts of the fitness function.
These algorithms are however generic and do not address the issue of selecting  learning rate for ML training with DP gradients and forecasting the quality of the trained ML model based on dataset sizes and privacy budgets. Forecasting the performance of privacy-preserving ML algorithms can be used to understand the value of collaboration between distributed private datasets. Without such forecasts the private data owners might need to forgo their private datasets so that a trusted third-party can compare the performance of the private ML model with the ML model trained in absence of privacy concerns (as otherwise there is no ground truth for comparison in general). Asynchronous optimization has been also utilized in the past for ML purposes; see, e.g.,~\cite{mnih2016asynchronous,smyth2009asynchronous,mcmahan2014delay}. These studies however do not consider additive DP noises and their impact on quality of trained ML models.



\section{Asynchronous ML Training with DP}\label{sec:DistributedML}

We consider $N\in\mathbb{N}$ private data owners connected to a central learning node, referred to as learner, responsible for training a ML model.

\begin{figure}[t!]
    \centering
\begin{tikzpicture}[scale=0.8]
\node[scale=0.8,draw,thick,align=center,color=black,shape=document,minimum width=12mm,minimum height=14mm,shape=document,inner sep=2ex] (1) at (-3,0.5) {};
\node[] at (-3,0.5) {
\begin{minipage}[t]{1cm}
\centering \scriptsize 
Data \\ owner 1
\end{minipage}
};
\node[] at (-5,0.5) {
\begin{minipage}{3cm}
\scriptsize $\mathcal{D}_1=\{(x_i,y_i)\}_{i=1}^{n_1}$ \\
$\epsilon_1$-DP response
\end{minipage}
};
\node[scale=0.8,draw,thick,align=center,color=black,shape=document,minimum width=12mm,minimum height=14mm,shape=document,inner sep=2ex] (2) at (-3,-1.2) {};
\node[] at (-3,-1.2) {
\begin{minipage}[t]{1cm}
\centering \scriptsize 
Data \\ owner 2
\end{minipage}
};
\node[] at (-5,-1.2) {
\begin{minipage}{3cm}
\scriptsize $\mathcal{D}_2=\{(x_i,y_i)\}_{i=1}^{n_2}$ \\
$\epsilon_2$-DP response
\end{minipage}
};
\node[scale=0.8,draw,thick,align=center,color=black,shape=document,minimum width=12mm,minimum height=14mm,shape=document,inner sep=2ex] (3) at (-3,-2.9) {};
\node[] at (-3,-2.9) {
\begin{minipage}[t]{1cm}
\centering \scriptsize 
Data \\ owner 3
\end{minipage}
};
\node[] at (-5,-2.9) {
\begin{minipage}{3cm}
\scriptsize $\mathcal{D}_3=\{(x_i,y_i)\}_{i=1}^{n_3}$ \\
$\epsilon_3$-DP response
\end{minipage}
};
\node[] at (-3,-4.4) {\Large .};
\node[] at (-3,-4.2) {\Large .};
\node[] at (-3,-4.0) {\Large .};
\node[scale=0.8,draw,thick,align=center,color=black,shape=document,minimum width=12mm,minimum height=14mm,shape=document,inner sep=2ex] (4) at (-3,-5.5) {};
\node[] at (-3,-5.5) {
\begin{minipage}[t]{1cm}
\centering \scriptsize 
Data \\ owner N
\end{minipage}
};
\node[] at (-5,-5.5) {
\begin{minipage}{3cm}
\scriptsize $\mathcal{D}_N=\{(x_i,y_i)\}_{i=1}^{n_N}$ \\
$\epsilon_N$-DP response
\end{minipage}
};
\node[] (L) at (1,-2.2) {\includegraphics[width=1.8cm]{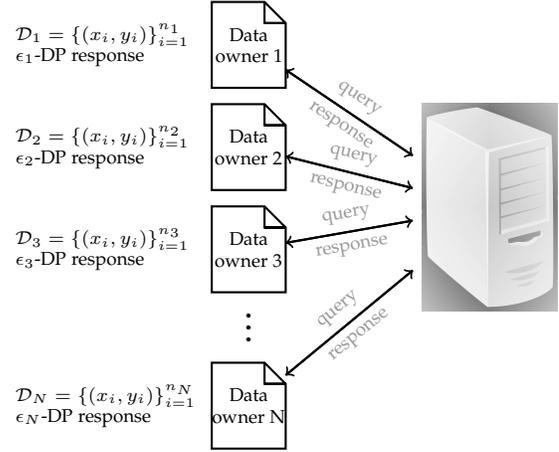}};
\draw [thick,->] (L) to [bend left=0] node [sloped,anchor=center,above]  {\scriptsize \color{black!40} query} (1);
\draw [thick,->] (1) to [bend right=0] node [sloped,anchor=center,below]  {\scriptsize \color{black!40} response} (L);
\draw [thick,->] (L) to [bend left=0] node [sloped,anchor=center,above]  {\scriptsize \color{black!40} query} (2);
\draw [thick,->] (2) to [bend right=0] node [sloped,anchor=center,below]  {\scriptsize \color{black!40} response} (L);
\draw [thick,->] (L) to [bend left=0] node [sloped,anchor=center,above]  {\scriptsize \color{black!40} query} (3);
\draw [thick,->] (3) to [bend right=0] node [sloped,anchor=center,below]  {\scriptsize \color{black!40} response} (L);
\draw [thick,->] (L) to [bend left=0] node [sloped,anchor=center,above]  {\scriptsize \color{black!40} query} (4);
\draw [thick,->] (4) to [bend right=0] node [sloped,anchor=center,below]  {\scriptsize \color{black!40} response} (L);
\end{tikzpicture}
    \caption{Communication structure between a central learner and multiple data owners with private datasets.}
    \label{fig:0}
\end{figure}

Figure~\ref{fig:0} depicts the communication structure between the learner and the private data owners. The set of data owners is denoted by $\mathcal{N}:=\{1,\dots,N\}$. The data owners possess a private training dataset composed of inputs $x_i$ and outputs $y_i$. The dataset is denoted by $\mathcal{D}_i:=\{(x_i,y_i)\}_{i=1}^{n_i}\subseteq \mathbb{X}\times\mathbb{Y} \subseteq\mathbb{R}^{p_x} \times\mathbb{R}^{p_y}$. 

%

Informally, an ML model is a meaningful relationship between inputs and outputs in a training dataset. The ML model is $\mathfrak{M}(\cdot;\theta)$ for some mapping $\mathfrak{M}:\mathbb{X}\times\mathbb{R}^{p_\theta}\rightarrow\mathbb{Y}$ with $\theta\in\mathbb{R}^{p_\theta}$ denoting the parameters of the ML model. The learner in Figure~\ref{fig:0} aims to train the ML model $\mathfrak{M}(\cdot;\theta)$ based on the available training datasets $\mathcal{D}_i$, $\forall i\in\mathcal{N}$, by solving the optimization problem in
\begin{align}\label{eqn:ML}
\theta^*\in\argmin_{\theta\in\Theta}f(\theta),
\end{align}
where $\Theta:=\{\theta\in \mathbb{R}^{p_\theta}\,|\,\| \theta \|_\infty\leq \theta_{\max}\}$ and $f:\mathbb{R}^{p_\theta}\rightarrow\mathbb{R}$ is the fitness for ML model parameter $\theta$, i.e., the fitness of ML model $\mathfrak{M}(\cdot;\theta)$ for relating the inputs and outputs in the training dataset $\cup_{j\in\mathcal{N}}\mathcal{D}_j$, given by
\begin{align}
f(\theta)
:=& \label{eqn:fitnessfunction}
g(\theta)+\frac{1}{n}\sum_{\{x,y\}\in\bigcup_{j\in\mathcal{N}}\mathcal{D}_j} \ell(\mathfrak{M}(x;\theta),y).
\end{align}  
In the fitness~\eqref{eqn:fitnessfunction},  $g(\theta)$ is a regularizing term, $\ell(\mathfrak{M}(x;\theta),y)$ is a loss function capturing the distance between the output of the ML model $\mathfrak{M}(x;\theta)$ and the true output $y$, and $n=\sum_{j\in\mathcal{N}}n_j$. Finally, note that we can select a large enough $\theta_{\max}$ so that, if desired, training on $\Theta$ does not add any conservatism (in comparison to the unconstrained case). 

In what follows, we present our ML learning algorithm for solving~\eqref{eqn:ML}. To do so, the learner must update the ML model based on the gradient of the fitness function~\eqref{eqn:fitnessfunction}. Noting that the learner might not have the communication and computational capacities required to interact with all the data owners at the same time, we consider the design constraint of having an asynchronous interaction between the data owners and the learner. The asynchronous setup implies that the learner can only communicate with one of the data owners at each given iteration and thus can only access the gradient of the part of the fitness that depends on the data possessed by the communicating data owner. This makes the task of updating the ML model challenging as the learner would not know the direction for the best model update. In fact, an update direction that is good for one data owner might not be good for all the others. To alleviate this problem, the learner updates the ML model with small, yet constant, learning rates. It also shows inertia in updating the ML model to avoid significant changes because of the gradient of just one data owner. The constant learning rate and the inertia of the learner allow the gradients of all the data owners to get mixed with across time so that the learner follows the direction for the best model update. In the remainder of this section, we clarify all the steps in the algorithm.

We model the internal clock of the data owner by Poisson point processes with rates of one. These clocks are not in any form synchronized and equal rate (of one) for the clocks simply implies that the data owners communicate with the learner with equal probability (not at equal times). At random times, the Poisson processes instigate communication between the data owners and the learner on a one-on-one basis. The Poisson process model is often utilized for analysis of asynchronous multi-agent systems~\cite{ram2009asynchronous,heidelberger1982queueing,lagunoff1997asynchronous}. Let the time instants in which the data owners communicate with the learner be given by
\begin{align*}
    0=t_1\leq t_2\leq \cdots \leq  t_k \leq \cdots \leq t_T.
\end{align*}
At each time instant $t_k$, $k\in\mathbb{N}$, one the data owners at random communicates with the learner. We use the notation $i_k\in\mathbb{N}$ to denote the index of that data owner that is communicating with the learner at time instance $t_k$. 

Two approaches can be utilized in the asynchronous communication. One approach is \textbf{broadcasting by the learner}. In this scenario, the learner, in regular time intervals, broadcasts gradient queries to all data owners (some might be listening while others not). Whenever one of the data owners responds, the index $k$ is incremented. Let $t_k$ denote the time at which the communication takes place and $i_k$ denote the index of the communicating data owner. Another approach is \textbf{requesting for update by the data owner}. In this scenario, the leaner is constantly listening for requests of update. Whenever a data owner submits a request, the index $k$ is incremented with $t_k$ denoting the time  and $i_k$ denoting the index of the data owner. At this point, the learner only communicates with that data owner until the update is over.

At each iteration, the learner submits a gradient query of the form
\begin{align} \label{eqn:gradient_query}
    \mathcal{Q}_{i_k}(\mathcal{D}_{i_k};\theta):= \frac{1}{n_{i_k}}\sum_{\{x,y\}\in\mathcal{D}_{i_k}} \nabla_\theta\ell(\mathfrak{M}(x;\theta),y) \in\mathcal{Q}
\end{align} 
to the communicating data owner $i_k\in\mathcal{N}$. Here, $\mathcal{Q}$ is the output space of the queries. The communicating data owner $i_k\in\mathcal{N}$ provides the DP response 
\begin{align}\label{eqn:gradient_DP_response}
    \overline{\mathcal{Q}}_{i_k}(\mathcal{D}_{i_k};\theta)=\mathcal{Q}_{i_k}(\mathcal{D}_{i_k};\theta)+w_{i_k}(k)
\end{align}
to the gradient query $\mathfrak{Q}_{i_k}(\mathcal{D}_{i_k};\theta)$. Here, $w_{i_k}(k)$ is a privacy-preserving additive noise to ensure DP.

\begin{definition}[Differential Privacy] \label{def:dp} Responses of data owner $\ell\in\mathcal{N}$ are $\epsilon_\ell$-differentially private (or $\epsilon_\ell$-DP) over the horizon $T$ if
\begin{align*}
\mathbb{P}\bigg\{(\overline{\mathfrak{Q}}_\ell(\mathcal{D}_\ell;k&))_{k:i_k=\ell}\in\mathcal{Y}\bigg\}\\[-.3em]
&\leq \exp(\epsilon_\ell)\mathbb{P}\bigg\{(\overline{\mathfrak{Q}}_\ell(\mathcal{D}'_\ell;k))_{k:i_k=\ell}\in\mathcal{Y}\bigg\},
\end{align*}
where $\mathcal{Y}$ is any Borel-measurable subset of $\mathcal{Q}^{|\{k:i_k=\ell\}|}$, and $\mathcal{D}_\ell$ and $\mathcal{D}'_\ell$ are two adjacent datasets differing at most in one entry, i.e., $|\mathcal{D}_\ell\setminus\mathcal{D}'_\ell|=|\mathcal{D}'_\ell\setminus\mathcal{D}_\ell|\leq 1$.
\end{definition}

We make the following standing assumptions throughout the paper for the purpose of theoretical analysis.

\begin{assumption} $g(\theta)$ is $\sigma$ strongly convex in $\theta$ and $\ell(\mathfrak{M}(x;\theta),y)$ is convex in~$\theta$.
\end{assumption}

\begin{assumption} \label{assum:gradientbound} The following properties hold:
\begin{enumerate}
\item $\Xi_g:=\sup_{\theta\in\Theta}\|\nabla_\theta g(\theta)\|_2<\infty$;
\item $\Xi:=\sup_{\theta\in\Theta}\sup_{(x,y)\in\mathbb{X}\times\mathbb{Y}}\|\nabla_\theta \ell(\mathfrak{M}(x;\theta),y)\|_2<\infty$.
\end{enumerate}
\end{assumption}

\begin{assumption} $T\in\mathbb{N}$ is the maximum number of iterations for communication between data owners and learner.
\end{assumption}

\begin{algorithm}[t!]
\caption{\label{alg:0} Asynchronous ML learning using DP gradients for strongly-convex smooth fitness cost.}
\begin{algorithmic}[1]
\REQUIRE $T\in\mathbb{N}$, $\rho\in\mathbb{R}_{\geq 0}$
\ENSURE $(\theta_{1,k},\theta_{2,k},\dots,\theta_{N,k},\theta_{L,k})_{k=1}^T$
\STATE Learner: Initialize $\theta_{1,0}=\cdots=\theta_{N,0}=\theta_{L,0}=0$
\FOR{$k=1,\dots,T$}
\STATE Randomly at uniform select data owner $i_k$
\STATE Learner: Compute $\bar{\theta}_{k}$ according to~\eqref{eqn:learners_hat_theta_k} 
\STATE Learner: Submit gradient query $\mathcal{Q}_{i_k}(\mathcal{D}_{i_k};\bar{\theta}_{k})$ to data owner $i_k$ according to~\eqref{eqn:gradient_query}
\STATE Data owner $i_k$: Provide DP response according to~\eqref{eqn:gradient_DP_response}
\STATE Learner: Update ML models according to~\eqref{eqn:learners_update_local} and~\eqref{eqn:learners_update_central}
\ENDFOR
\end{algorithmic}
\end{algorithm}

\begin{theorem} \label{tho:1} The policy of data owners in~line 6 of Algorithm~\ref{alg:0}  for responding to the queries over the horizon $\{1,\dots,T\}$  is  $\epsilon_i$-DP, $\forall i\in\mathcal{N}$, if $w_i(k)$ are statistically independent Laplace noises with  scale $2\Xi T/(n_i \epsilon_i)$. 
\end{theorem}

\begin{proof} Due to space constraints, the proofs are presented as supplementary material.
\end{proof}

We consider an approach in which the leaner keeps track of a central ML model, i.e., $\theta_{L,k}$, and $N$ copies of it for each data owners, i.e., $\theta_{i,k}$ for each $i=1,\dots,N$. This is motivated by the algorithm in~\cite{5454103} that forms the basis of our ML training algorithm. The local copies are only updated when the corresponding data owner is communicating with the learner. This is to keep track of the updates for each data owner. The update for the local ML model is given by
\begin{align} \label{eqn:learners_update_local}
    \theta_{i_k,k}=&\Pi_\Theta\Bigg[\bar{\theta}_{k}-\frac{N\rho }{T^2\sigma}\Bigg(\frac{1}{2N}\nabla_\theta g(\bar{\theta}_{k})\nonumber\\
    &\hspace{1in}+\frac{n_{i_k}}{n}\overline{\mathcal{Q}}_{i_k}(\mathcal{D}_{i_k};\bar{\theta}_{k})\Bigg) \Bigg],
\end{align}
where
\begin{align} \label{eqn:learners_hat_theta_k}
    \bar{\theta}_{k}=\frac{1}{2}(\theta_{L,k-1}+\theta_{i_k,k-1}).
\end{align}
Note that the learner updates the ML model with small, yet constant, learning rates. The learner also shows inertia in updating the central ML model so that it does not change the model significantly because of the gradient of just one data owner. The update for the central ML model is given by
\begin{align} \label{eqn:learners_update_central}
\theta_{L,k}=&\Pi_\Theta\left[\bar{\theta}_{k}-\frac{(N-1)\rho}{NT^2\sigma}\nabla_\theta g(\bar{\theta}_{k}) \right].
\end{align}
The constant learning rate and the inertia of the learner allow the gradients of all the data owners to get mixed with each other across time so that the learner follow the direction for the best model update. All the steps of the learner and the data owners for generating queries, responding to the queries, and using the DP responses for updating the ML model are summarized in Algorithm~\ref{alg:0}. Finally, note that in step 3 of  Algorithm~\ref{alg:0} it states that the data owners are selected uniformly at random. This is compatible with the Poisson process clocks. The first data owner whose Poisson clock ticks communicates with the learner and because of the symmetry this happens with equal probability between the data owners (hence, in each iteration, one of the agents with uniform probability communicates with the learner).

\section{Performance of Private ML Models} \label{sec:theory}
For Algorithm~\ref{alg:0}, we can prove the following convergence result under the assumptions of strong convexity and smoothness of the ML fitness function.

\begin{theorem} \label{tho:2} For any $N$, there exist constants\footnote{See the proof of the theorem in the supplementary materials for the exact constants.} $c_1,c_2,c'_1,c'_2>0$ such that the iterates of Algorithm~\ref{alg:0} satisfy
\begin{align}
    \mathbb{E}\{\|\theta_{L,T}-\theta^*\|_2^2\}
    \hspace{-.03in}\leq& c_1\hspace{-.03in}\sqrt{\hspace{-.03in}\frac{1}{T^2}\hspace{-.03in}+\hspace{-.03in}N\hspace{-.03in}\sum_{i\in\mathcal{N}}\hspace{-.03in}\left(
    \frac{1}{T}\hspace{-.03in}+\hspace{-.03in}\frac{2\sqrt{2}}{n\epsilon_i}\right)^{\hspace{-.03in}2}}
\nonumber    
    \\[-.2em]
     &+\hspace{-.03in}c_2\hspace{-.03in}\left(\hspace{-.03in}\frac{1}{T^2}\hspace{-.03in}+\hspace{-.03in}N\hspace{-.03in}\sum_{i\in\mathcal{N}}\hspace{-.03in}\left(
    \frac{1}{T}\hspace{-.03in}+\hspace{-.03in}\frac{2\sqrt{2}}{n\epsilon_i}\right)^{\hspace{-.05in}2}\right)\hspace{-.03in}.
    \label{eqn:bound:1}
\end{align}
and
\begin{align}
    \mathbb{E}\{f(\theta_{L,T})\}\hspace{-.03in}-\hspace{-.03in}f(\theta^*)
    \hspace{-.03in}\leq &c'_1\hspace{-.03in}\sqrt{\hspace{-.03in}\frac{1}{T^2}\hspace{-.03in}+\hspace{-.03in}N\hspace{-.03in}\sum_{i\in\mathcal{N}}\hspace{-.03in}\left(\hspace{-.03in}
    \frac{1}{T}\hspace{-.03in}+\hspace{-.03in}\frac{2\sqrt{2}}{n\epsilon_i}\right)^{\hspace{-.03in}2}}
\nonumber    
    \\[-.2em]
     &+\hspace{-.03in}c'_2\hspace{-.03in}\left(\hspace{-.03in}\frac{1}{T^2}\hspace{-.03in}+\hspace{-.03in}N\hspace{-.03in}\sum_{i\in\mathcal{N}}\hspace{-.03in}\left(\hspace{-.03in}
    \frac{1}{T}\hspace{-.03in}+\hspace{-.03in}\frac{2\sqrt{2}}{n\epsilon_i}\right)^{\hspace{-.05in}2}\right)\hspace{-.03in}.
    \label{eqn:bound:2}
\end{align}
\end{theorem}

\begin{proof} Due to space constraints, the proofs are presented as supplementary material.
\end{proof}

For large enough learning horizon $T$, the upper bound~\eqref{eqn:bound:1} takes the form of 
\begin{align} \label{eqn:1_limiting}
    \mathbb{E}\{\|\theta_{L,T}-\theta^*\|_2^2\}
    \leq& \frac{\bar{c}_1}{n}\sqrt{\sum_{i\in\mathcal{N}}\frac{1}{\epsilon_i^2}}+\frac{\bar{c}_2}{n^2}\left(\sum_{i\in\mathcal{N}}\frac{1}{\epsilon_i^2}\right),
\end{align}
where $\bar{c}_1=\sqrt{8N}c_1$ and $\bar{c}_2=8Nc_2$. Similarly, for large $T$, the upper bound~\eqref{eqn:bound:1} takes the form of 
\begin{align} \label{eqn:2_limiting}
    \mathbb{E}\{f(\theta_{L,T})\}\hspace{-.03in}-\hspace{-.03in}f(\theta^*)
    \leq& \frac{\bar{c}'_1}{n}\sqrt{\sum_{i\in\mathcal{N}}\frac{1}{\epsilon_i^2}}+\frac{\bar{c}'_2}{n^2}\left(\sum_{i\in\mathcal{N}}\frac{1}{\epsilon_i^2}\right),
\end{align}
where again $\bar{c}'_1=\sqrt{8N}c'_1$ and $\bar{c}'_2=8Nc'_2$. This takes the form of the performance bound in~\cite{farokhi_ieee_sp_2020}. Under the assumption that all the data owners have equal privacy budgets $\epsilon_i=\epsilon$, $\forall i$, the bound in~\eqref{eqn:2_limiting} scales as $\epsilon^{-2}$. This bound matches the lower and upper bounds in~\cite{bassily2014private} for strongly convex loss functions. The same outcome also holds if $N=1$ and $\epsilon_1=\epsilon$ which captures centralized privacy-preserving learning. 

We can introduce the cost of privacy (CoP) as the difference of the fitness for privacy-preserving ML model and the fitness for trained ML model in the absence of privacy concerns. The inequalities in~\eqref{eqn:1_limiting} and \eqref{eqn:2_limiting} show that CoP is  inversely proportional to the combined size of the  training datasets squared and the sum of the privacy budgets squared. 

\section{Experimental Validation} \label{sec:numerical}

In this section, we investigate the performance of Algorithm~\ref{alg:0} on real datasets from the financial and health domains. In our experiments, the datasets have significantly different sizes and the size of the training datasets influence the performance of both non-private and private ML models. Hence, we factor out the effects of the size of the training datasets on the performance of the learning by only considering the relative fitness, defined as $\psi(\theta):=f(\theta)/f(\theta^*)-1.$ This measure captures the quality of any ML model $\theta$  in comparison to the  non-private ML model $\theta^*$ in terms of the fitness in~\eqref{eqn:fitnessfunction}. By definition, $\psi(\theta)\geq 0$ for any ML model $\theta$. The larger $\psi(\theta)$, the worse the performance of ML model $\theta$ in comparison with the non-private ML model $\theta^*$. 

Datasets and codes for developing the experimental results reported in this paper are available in an online Git repository~\cite{proofs_of_the_paper}. 

\begin{figure}[t]
\centering
\begin{tabular}{c}
\begin{tikzpicture}
\node[] at (0,0) {
\includegraphics[width=.85\linewidth]{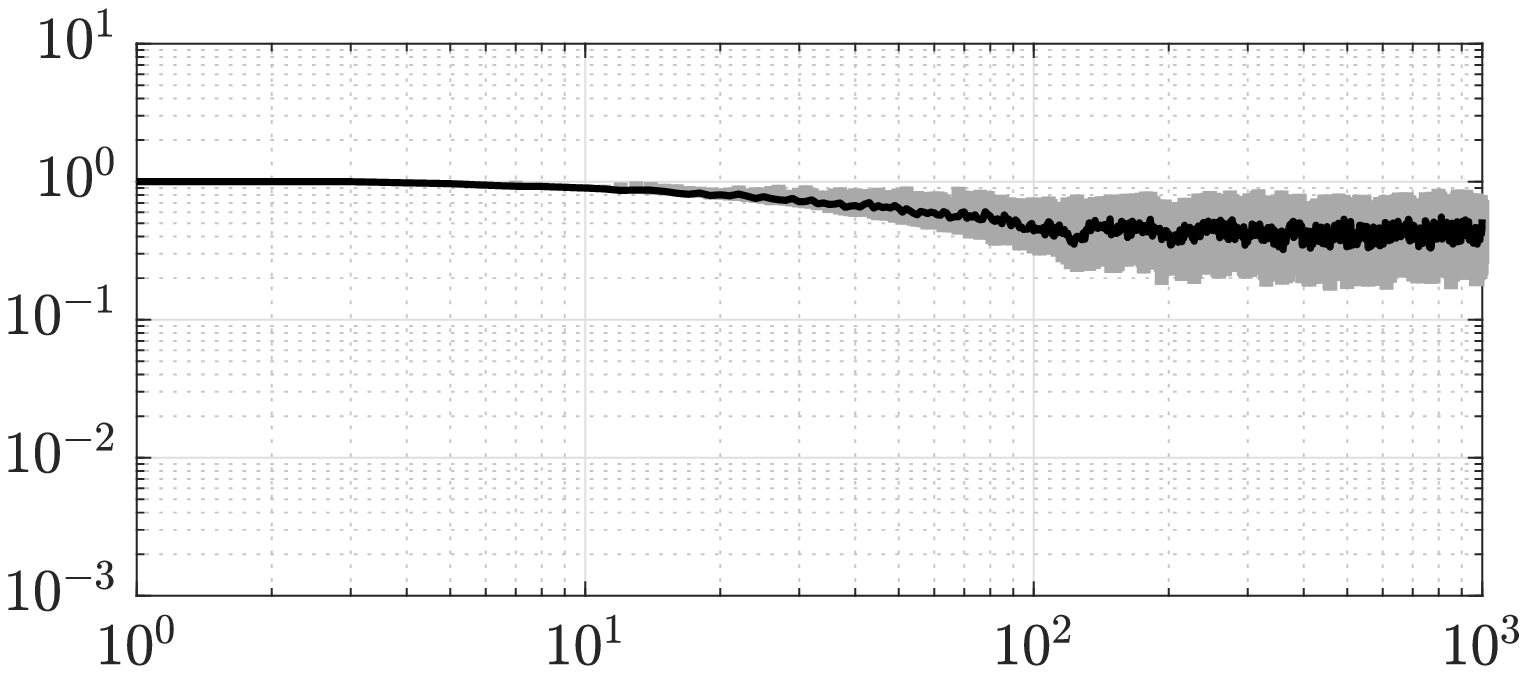}};
\node[] at (0,-1.5) {\footnotesize $k$};
\node[rotate=90] at (-3.6,0) {\footnotesize $\psi(\theta)$};
\node[draw,fill=white,minimum width=3.1cm] at (-1.0,-0.7) {$\epsilon_1=\epsilon_2=\epsilon_3=0.1$};
\end{tikzpicture}
\\[-.6em]
\begin{tikzpicture}
\node[] at (0,0) {
\includegraphics[width=.85\linewidth]{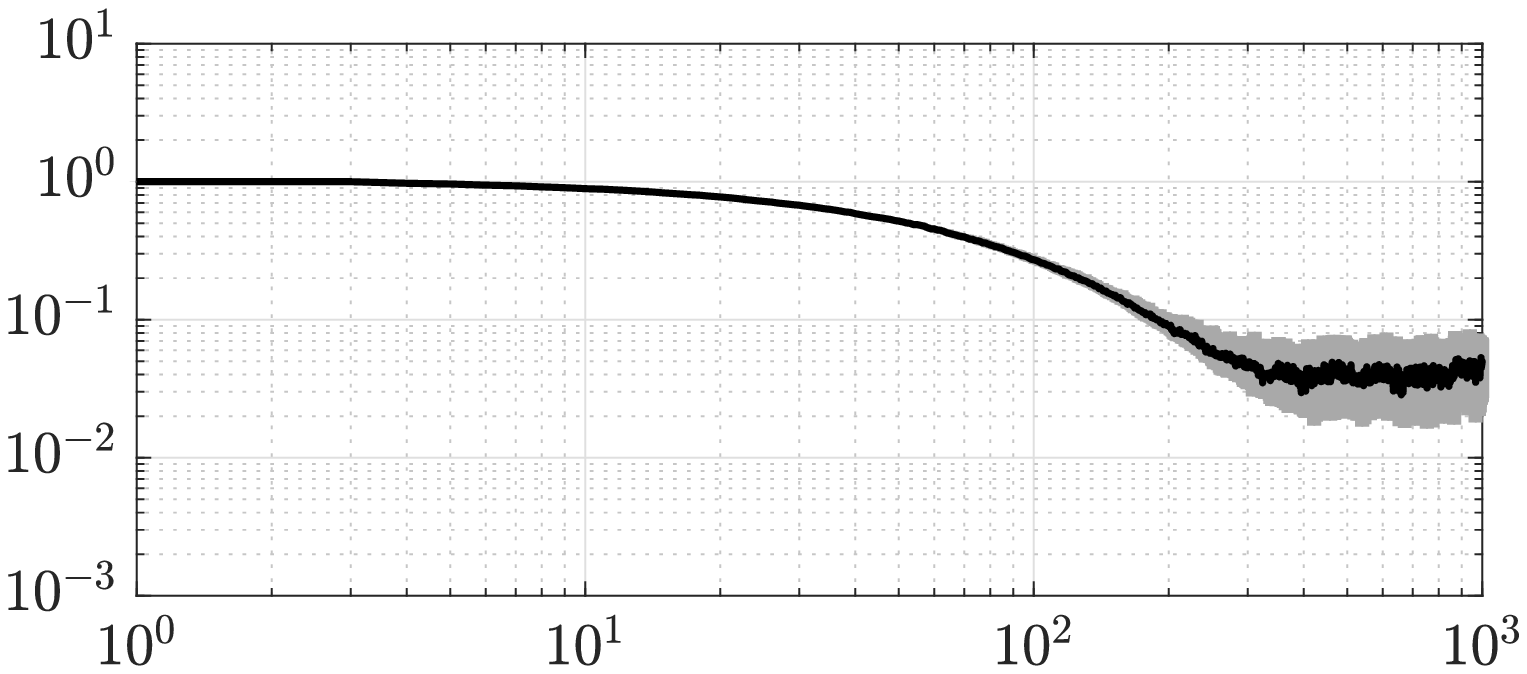}};
\node[] at (0,-1.5) {\footnotesize $k$};
\node[rotate=90] at (-3.6,0) {\footnotesize $\psi(\theta)$};
\node[draw,fill=white,minimum width=3.1cm] at (-1.0,-0.7) {$\epsilon_1=\epsilon_2=\epsilon_3=1.0$};
\end{tikzpicture}
\\[-.6em]
\begin{tikzpicture}
\node[] at (0,0) {
\includegraphics[width=.85\linewidth]{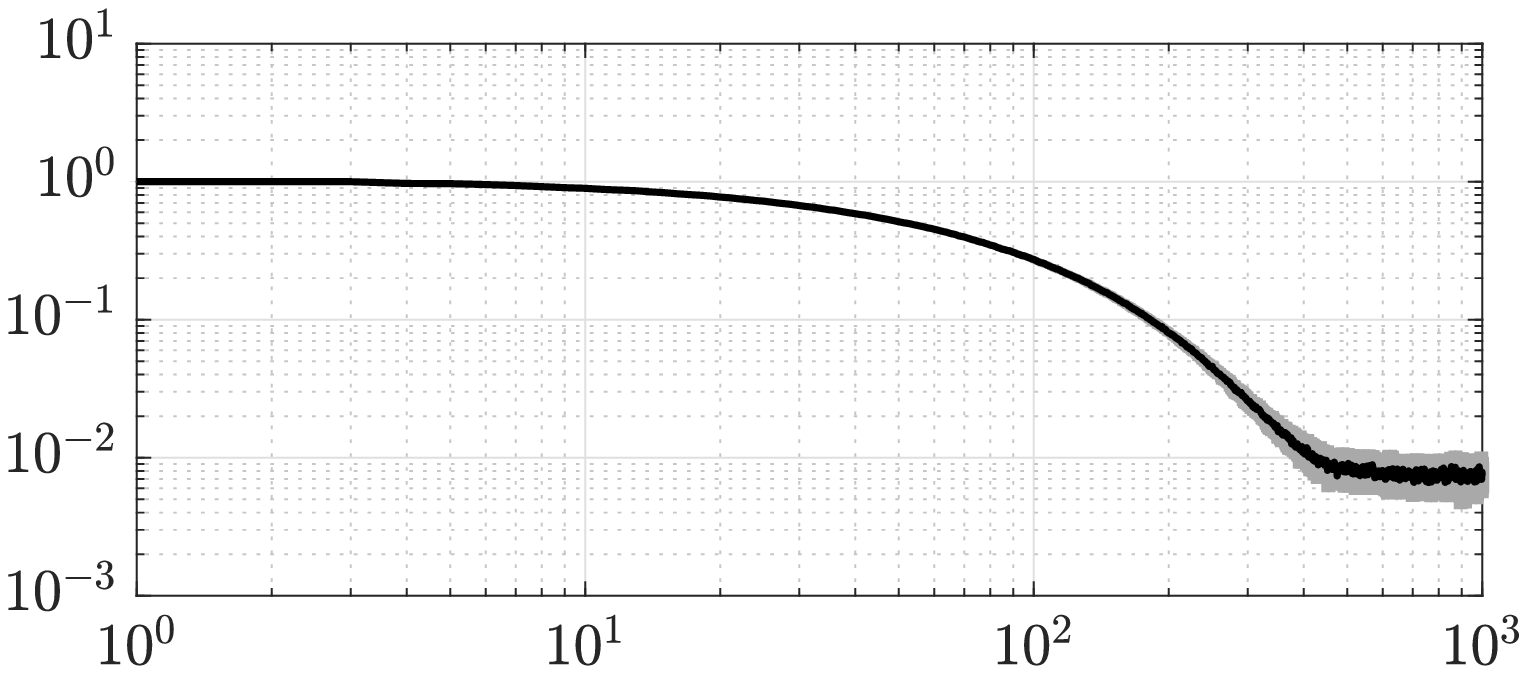}};
\node[] at (0,-1.5) {\footnotesize $k$};
\node[rotate=90] at (-3.6,0) {\footnotesize $\psi(\theta)$};
\node[draw,fill=white,minimum width=3.1cm] at (-1.0,-0.7) {$\epsilon_1=\epsilon_2=\epsilon_3=10$};
\end{tikzpicture}
\end{tabular}
\caption{
\label{fig:a0}  
Percentile statistics of relative fitness of 100 runs of Algorithm~\ref{alg:0} for learning lending-interest-rates versus the iteration number $k$ for a learning horizon of $T=1,000$ iterations with three choices of privacy budgets $\epsilon_1=\epsilon_2=\epsilon_3$. The gray area illustrates the range of 25\% to 75\% percentiles and the black line shows the median of relative fitness. }
\end{figure}

\begin{figure*}
\centering
\begin{tikzpicture}
\node[] at (0,0) {\includegraphics[width=0.98\linewidth]{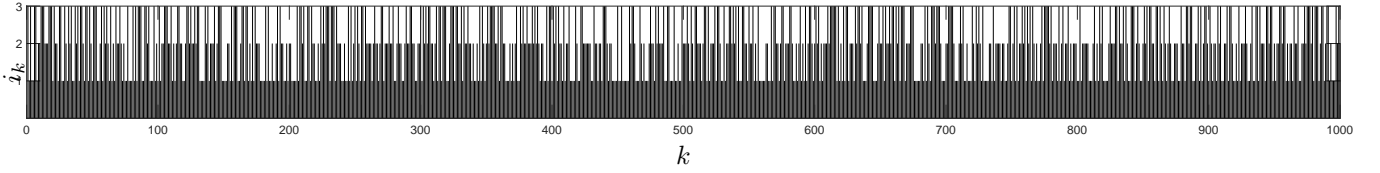}};
\node[] at (0,-1.1) {$k$};
\node[rotate=90] at (-8.9,0) {$i_k$};
\end{tikzpicture}
\vspace{-.3in}
\caption{
\label{fig:a00}
Example of communication timing for the asynchronous learning in Algorithm~\ref{alg:0} for learning lending-interest-rates, illustrating $i_k$ versus the iteration number $k$.
}
\end{figure*}

\begin{figure}[t]
\centering
\begin{tikzpicture}
\node[] at (0,0) {
\includegraphics[width=0.85\columnwidth]{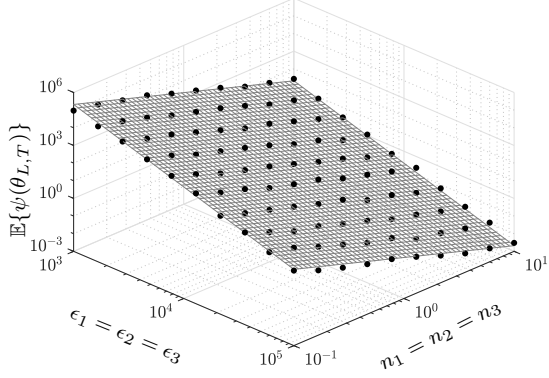}};
\node[rotate=27] at (+2.1,-2.1) {\footnotesize $n_1=n_2=n_3$};
\node[rotate=-22] at (-2.1,-2.1) {\footnotesize $\epsilon_1=\epsilon_2=\epsilon_3$};
\node[rotate=90] at (-3.5,0) {\footnotesize $\mathbb{E}\{\psi(\theta_{L,T})\}$};
\end{tikzpicture}
\caption{
\label{fig:a1} Relative fitness of Algorithm~\ref{alg:0} for learning lending-interest-rates after $T=1,000$ iterations versus the size of the datasets $n_1=n_2=n_3$ and the privacy budgets $\epsilon_1=\epsilon_2=\epsilon_3$. The mesh surface illustrates the bound in~\eqref{eqn:2_limiting} with $\bar{c}'_1=0$ and $\bar{c}'_2=2.1\times 10^9$. }
\end{figure}

\begin{figure}[t]
\centering
\begin{tabular}{c}
\begin{tikzpicture}
\node[] at (0,0) {
\includegraphics[width=.9\columnwidth]{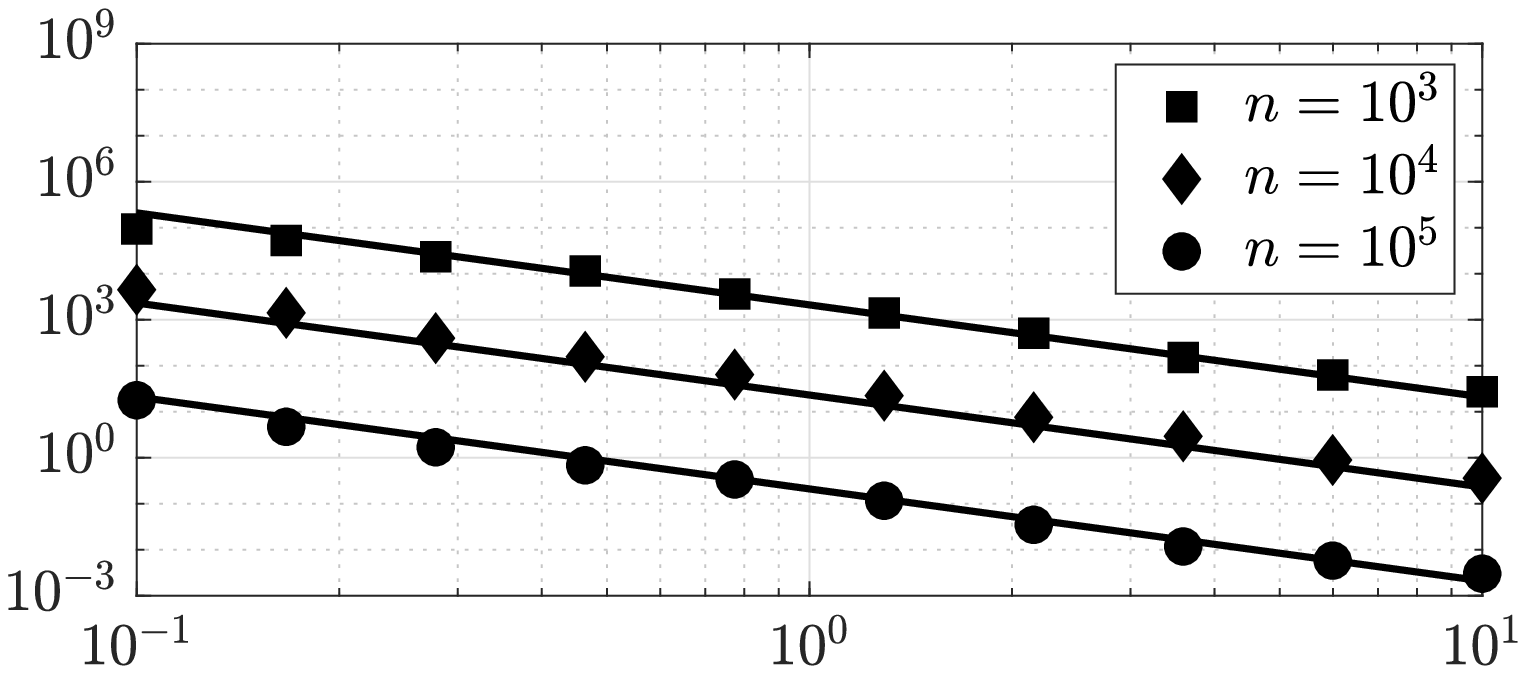}};
\node[] at (0,-1.8) {\footnotesize $\epsilon_1=\epsilon_2=\epsilon_3$};
\node[rotate=90] at (-3.6,0) {\footnotesize $\mathbb{E}\{\psi(\theta_{L,T})\}$};
\end{tikzpicture}
\\[-.7em]
\begin{tikzpicture}
\node[] at (0,0) {
\includegraphics[width=.9\columnwidth]{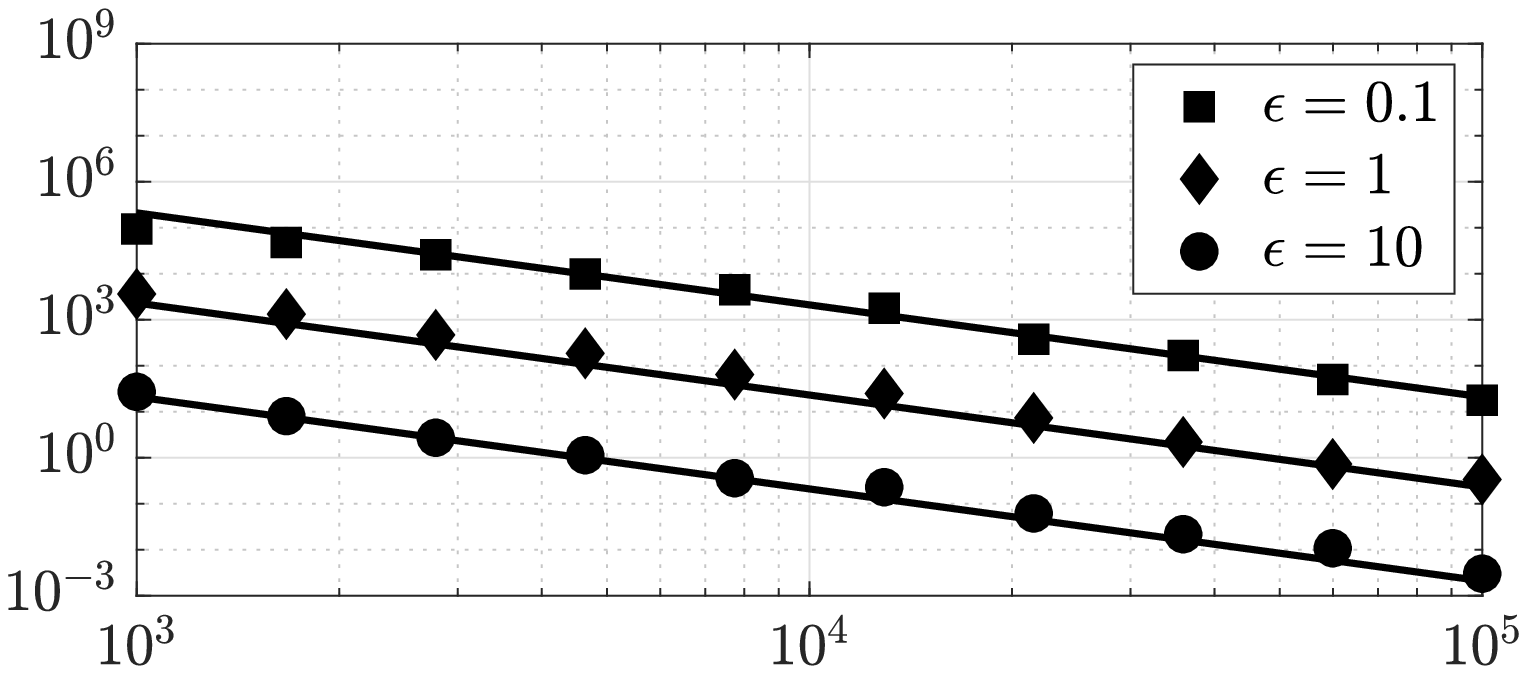}};
\node[] at (0,-1.8) {\footnotesize $n_1=n_2=n_3$};
\node[rotate=90] at (-3.6,0) {\footnotesize $\mathbb{E}\{\psi(\theta_{L,T})\}$};
\end{tikzpicture}
\end{tabular}
\caption{\label{fig:a3} Relative fitness of Algorithm~\ref{alg:0} for learning lending-interest-rates after $T=1,000$ iterations versus the privacy budget [top] and the size of the datasets [bottom]. The solid line illustrates the bound in~\eqref{eqn:2_limiting} with $\bar{c}'_1=0$ and $\bar{c}'_2=2.1\times 10^9$. }
\end{figure}

\subsection{Lending Dataset (Financial)}
We first train a linear regression model on lending datasets as an example of automating banking processes requiring access to sensitive private datasets.

\subsubsection{Dataset Description and Pre-Processing} We use a dataset  of anonymized loan application information from roughly 890,000 individuals~\cite{kaggle1}. We remove  unique identifiers, such as id and member id, and irrelevant attributes, such as URL addresses. We endeavour to train a linear regression model on this dataset. The input to the regression model are loan information, such as loan size, and applicant information, such as credit rating, state of residence and age. The model estimates the  annual interest rate for the loans. We encode categorical attributes, such as state of residence and loan grade, with integer numbers. 

In order to improve the numerical stability of the algorithm, we use Principal Component Analysis (PCA) to perform feature selection. We select the top ten important features. For this step, we only use the last ten-thousand entries of the dataset. We can assume that these entries are known to the learner and thus do not violate the distributed nature of the algorithm. This would have been a restrictive assumption if the learner used the entire dataset for the PCA (because the data owners must have agreed to perform PCA in collaboration without privacy concerns, which is contradiction with their original interest for privacy-preserving ML). Using the PCA, the learner can construct a dictionary for feature selection and communicate it to private data owners. 

\begin{figure}[t]
\centering
\begin{tikzpicture}
\node[] at (0,0) {
\includegraphics[width=.85\columnwidth]{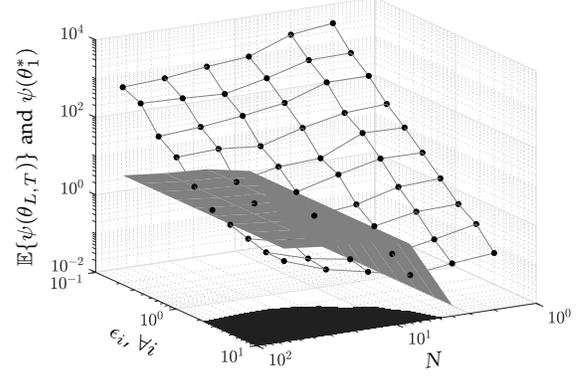}};
\node[rotate=90] at (-3.7,.3) {\footnotesize $\mathbb{E}\{\psi(\theta_{L,T})\}$ and $\psi(\theta_1^*)$};
\node[rotate=9] at (+1.7,-2.4) {\footnotesize $N$};
\node[rotate=-25] at (-2.3,-2.15) {\footnotesize $\epsilon_i$, $\forall i$};
\end{tikzpicture}
\caption{\label{fig:a4} Relative fitness of Algorithm~\ref{alg:0} for learning lending-interest-rates after $T=1,000$ iterations, $\mathbb{E}\{\psi(\theta_{L,T})\}$, versus the privacy budgets  $\epsilon_i$, $\forall i$, and the number of collaborating data owners $N$. The solid gray surface shows the relative fitness of the non-private ML model $\theta^*_1$, $\psi(\theta_1^*)$, constructed based on only the private data of the first data owner. If the relative fitness of Algorithm~\ref{alg:0} is smaller than the relative fitness of the non-private ML model $\theta^*_1$, collaboration benefits the first data owner (illustrated by the black region at the bottom of the figure).
}
\end{figure}

\subsubsection{Experiment Setup and Results} 

We start with an experiment evaluating the outcome of collaborations between $N=3$ banks. We use the linear regression model $y=\mathfrak{M}(x;\theta):=\theta^\top x$ with $\theta$ denoting the model parameters. The fitness function is given by $g_2(\mathfrak{M}(x;\theta),y)= \|y-\mathfrak{M}(x;\theta)\|_2^2$, and $g_1(\theta)=10^{-5}\theta^\top\theta$. The first data owner is assumed to possess the first $n_1$ entries of the dataset. The second data owner owns entries ranging from $n_1+1$ to $n_1+n_2$. Finally, the third data owner has access to  entries between $n_1+n_2+1$ to $n_1+n_2+n_3$ as its private dataset.

We start with demonstrating the convergence of Algorithm~\ref{alg:0} when $n_1=n_2=n_3=250,000$. Figure~\ref{fig:a0} illustrates the percentile statistics of the relative fitness $\psi(\theta_{L,k})$ for 100 runs of Algorithm~\ref{alg:0} versus the iteration number $k$ for the learning horizon $T=1,000$. Note that, in Algorithm~\ref{alg:0}, only one of the data owners communicates with the learner in each iteration. Figure~\ref{fig:a00} illustrates an example of communication timing for the asynchronous learning in Algorithm~\ref{alg:0}, illustrating $i_k$ versus the iteration number $k$. Recalling the stochastic nature of the algorithm, due to the DP noise in query responses, the relative fitness varies for each run of the algorithm. The gray area in Figure~\ref{fig:a0} shows 25\%--75\% percentiles of the relative fitness. The black solid lines in Figure~\ref{fig:a0} shows the median of relative fitness versus the iteration number. The median decreases across time until the algorithm converges to a neighbourhood of the solution of~\eqref{eqn:ML}. The relative fitness of the trained model also improves as $\epsilon_1=\epsilon_2=\epsilon_3$ increases. Note that smaller privacy budgets also increase the variations in the relative fitness (i.e., larger gray area).

Figure~\ref{fig:a1} illustrates the average relative fitness of the trained ML model  using Algorithm~\ref{alg:0} after $T=1,000$ iterations, $\mathbb{E}\{\psi(\theta_{L,T})\}$, versus the size of the private datasets $n_1=n_2=n_3$ and the privacy budgets $\epsilon_1=\epsilon_2=\epsilon_3$. The mesh surface shows the bound in~\eqref{eqn:2_limiting} with $\bar{c}'_1=0$ and $\bar{c}'_2=2.1\times 10^9$. This figure clearly shows the tightness of the result of Theorem~\ref{tho:2}. Note that, as expected, the relative fitness rapidly improves as the sizes of the datasets $n_1=n_2=n_3$ or the privacy budgets  $\epsilon_1=\epsilon_2=\epsilon_3$ increase.
 
Let us isolate the effects of the size of the datasets and the privacy budgets. Figure~\ref{fig:a3} shows the average relative fitness of the trained ML model  using Algorithm~\ref{alg:0} after $T=1,000$ iterations, $\mathbb{E}\{\psi(\theta_{L,T})\}$, versus the privacy budgets  $\epsilon_1=\epsilon_2=\epsilon_3$ [top] and the size of the datasets $n_1=n_2=n_3$ [bottom]. In this figure, the markers (i.e., {\small$\blacksquare$}, $\blacklozenge$, and \tikz\draw[black,fill=black] (0,0) circle (.6ex);) are from the experiments and the solid show the bound in~\eqref{eqn:2_limiting}. For both these cases, the bounds in Theorem~\ref{tho:2} are tight fits. Therefore, the theoretical results in Theorem~\ref{tho:2} match the experiments. 

Let us also demonstrate the value of collaboration between among many banks. Consider an experiment with $N$ banks each with $n_i=10,000$ records collaborating to train a regression model.  Figure~\ref{fig:a4} shows the average relative fitness of Algorithm~\ref{alg:0} for learning lending-interest-rates after $T=1,000$ iterations, $\mathbb{E}\{\psi(\theta_{L,T})\}$, versus the privacy budgets $\epsilon_i$, $\forall i$, and the number of the collaborating data owners $N$. The solid gray surface shows the relative fitness of the non-private ML model $\theta^*_1$, $\psi(\theta_1^*)$, constructed based on only the private data of the first data owner. Note that $\psi(\theta_1^*)$ is not random (as its construction does not require DP noise) and is not a function of $\epsilon_i$. If the relative fitness of Algorithm~\ref{alg:0} is smaller than the relative fitness of the non-private ML model $\theta^*_1$, collaboration benefits the first data owner, which is illustrated by the black region at the bottom of the figure. Evidently, the first data owner benefits from collaboration if there are more than 5 data owners with privacy budgets greater than or equal to 10 or if there are more than 100 data owners with privacy budgets greater than or equal to 2.5.

\subsection{Health-related Data}
Now, we use the hospital admission and discharge dataset from the New York State to validate the theoretical results. 

\subsubsection{Dataset Description and Pre-Processing} The dataset contains hospital visit and discharge information from nearly 2,350,000 de-identified  patients including information, such as characteristics, diagnoses, treatments, services, and charges. This dataset is made public by the Bureau of Health Informatics~\cite{nyhealthdata}. 
We train a linear regression model, as in the previous subsection, with inputs, such as age, gender, race, ethnicity, diagnosis code, procedure code, and drug code, to automatically determine the length of stay. This can be used as a tool for determining the capacity of hospitals in the future based on currently admitted patients. Similarly, we encode categorical attributes, such as gender and ethnicity, with integer numbers. We also remove attributes, such as total charges and costs,  as well as irrelevant attributes, such as the postcode. Similar to the lending data, in order to improve the numerical stability of the algorithm, we perform the PCA to  balance the features. We do so based on the last fifty-thousand entries of the dataset to ensure that the feature selection does not violate the distributed nature of the algorithm. 

\subsubsection{Experiment Setup and Results}
The data in~\cite{nyhealthdata} is tagged by the hospital name and code. There are 213 hospitals in the dataset. We focus on 86 hospital with at least 10,000 records. Experiments on the convergence of the algorithms and the tightness of theoretical bounds are similar to the lending data and are therefore eliminated due to space constraints. They are however presented as supplementary material.

\begin{figure}[t]
\centering
\begin{tikzpicture}
\node[] at (0,0) {
\includegraphics[width=.85\columnwidth]{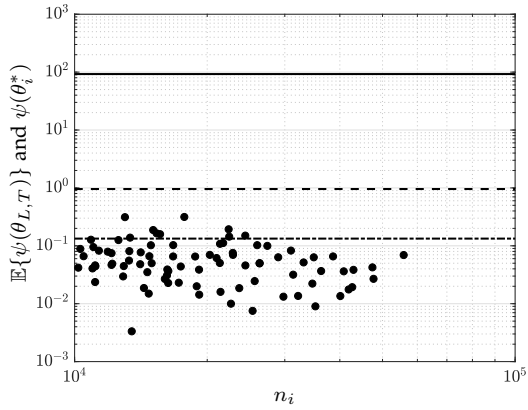}};
\node[rotate=90] at (-3.5,.3) {\footnotesize $\mathbb{E}\{\psi(\theta_{L,T})\}$ and $\psi(\theta_i^*)$};
\node[] at (0,-2.7) {\footnotesize $n_i$};
\end{tikzpicture}
\caption{\label{fig:b2} Relative fitness of Algorithm~\ref{alg:0} for learning  length of stay at hospital after $T=1,000$ iterations, $\mathbb{E}\{\psi(\theta_{L,T})\}$, for three choices of privacy budgets $\epsilon_i=0.1$ (black line),  $\epsilon_i=1$ (dashed line),  $\epsilon_i=10$ (dash-dotted line). The markers show the relative fitness of the non-private ML model $\theta^*_i$, $\psi(\theta_i^*)$, constructed based on only the private data of the $i$-th data owner versus the size of the data set owned by the $i$-th data owner. For $\epsilon=10$, eight hospitals benefit from collaboration. The relative fitness of the non-private ML model $\theta^*_i$ for these eight hospitals are above the dash-dotted line.  }
\end{figure}

\begin{figure}[t]
	\centering
	\begin{tabular}{c}
		\begin{tikzpicture}
		\node[] at (0,0) {
			\includegraphics[width=1\linewidth]{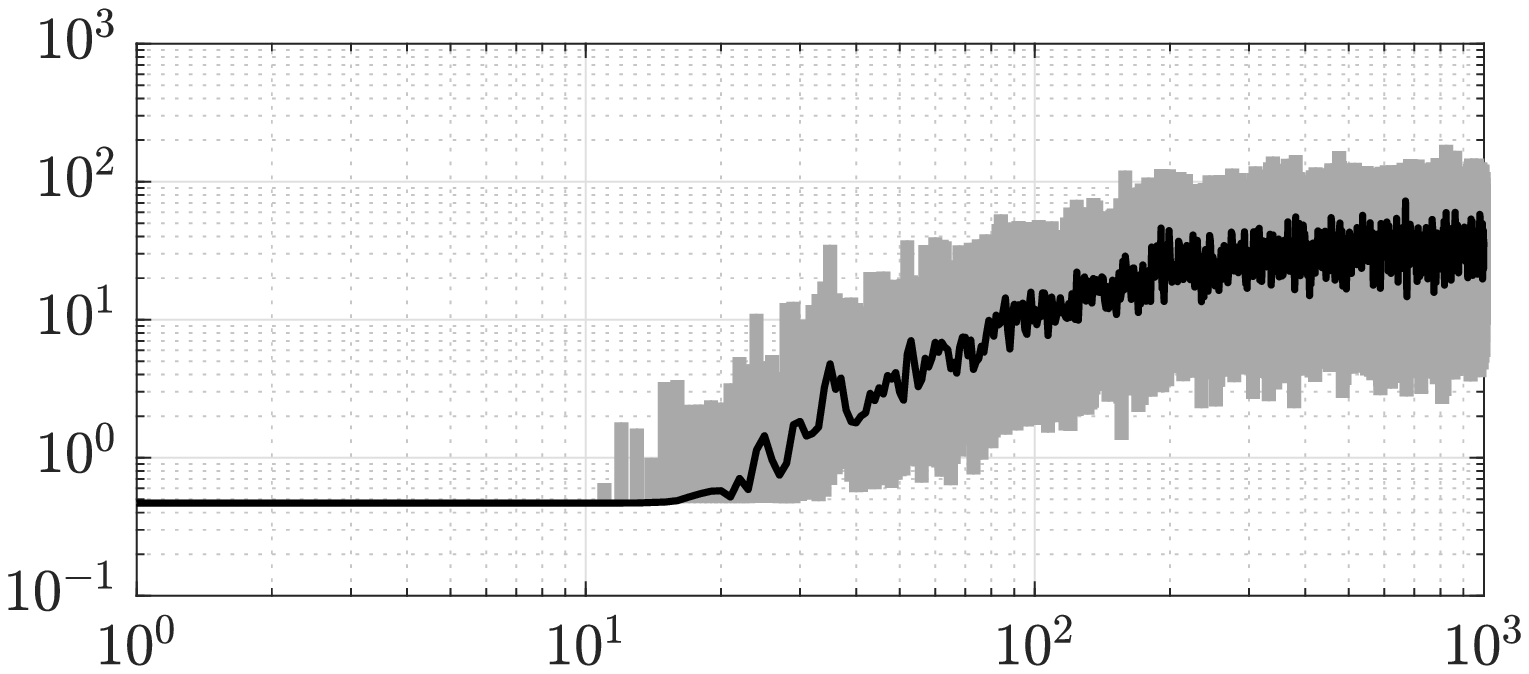}};
		\node[] at (0,-1.9) {\footnotesize $k$};
		\node[rotate=90] at (-4.3,0) {\footnotesize $\psi(\theta)$};
		\node[draw,fill=white,minimum width=3.1cm] at (-1.5,+1.1) {$\epsilon_1=\epsilon_2=\epsilon_3=0.1$};
		\end{tikzpicture}
		\\[-.5em]
		\begin{tikzpicture}
		\node[] at (0,0) {
			\includegraphics[width=1\linewidth]{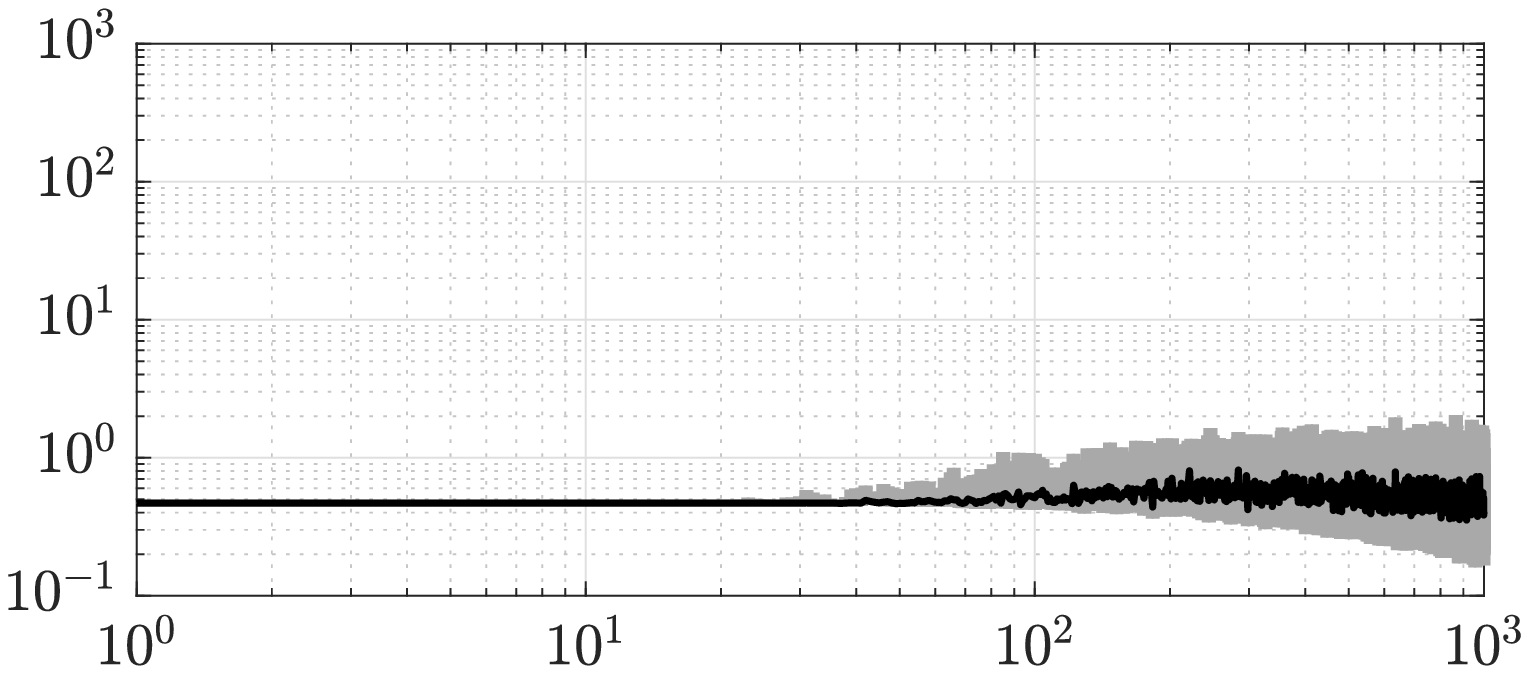}};
		\node[] at (0,-1.9) {\footnotesize $k$};
		\node[rotate=90] at (-4.3,0) {\footnotesize $\psi(\theta)$};
		\node[draw,fill=white,minimum width=3.1cm] at (-1.5,+1.1) {$\epsilon_1=\epsilon_2=\epsilon_3=1.0$};
		\end{tikzpicture}
		\\[-.5em]
		\begin{tikzpicture}
		\node[] at (0,0) {
			\includegraphics[width=1\linewidth]{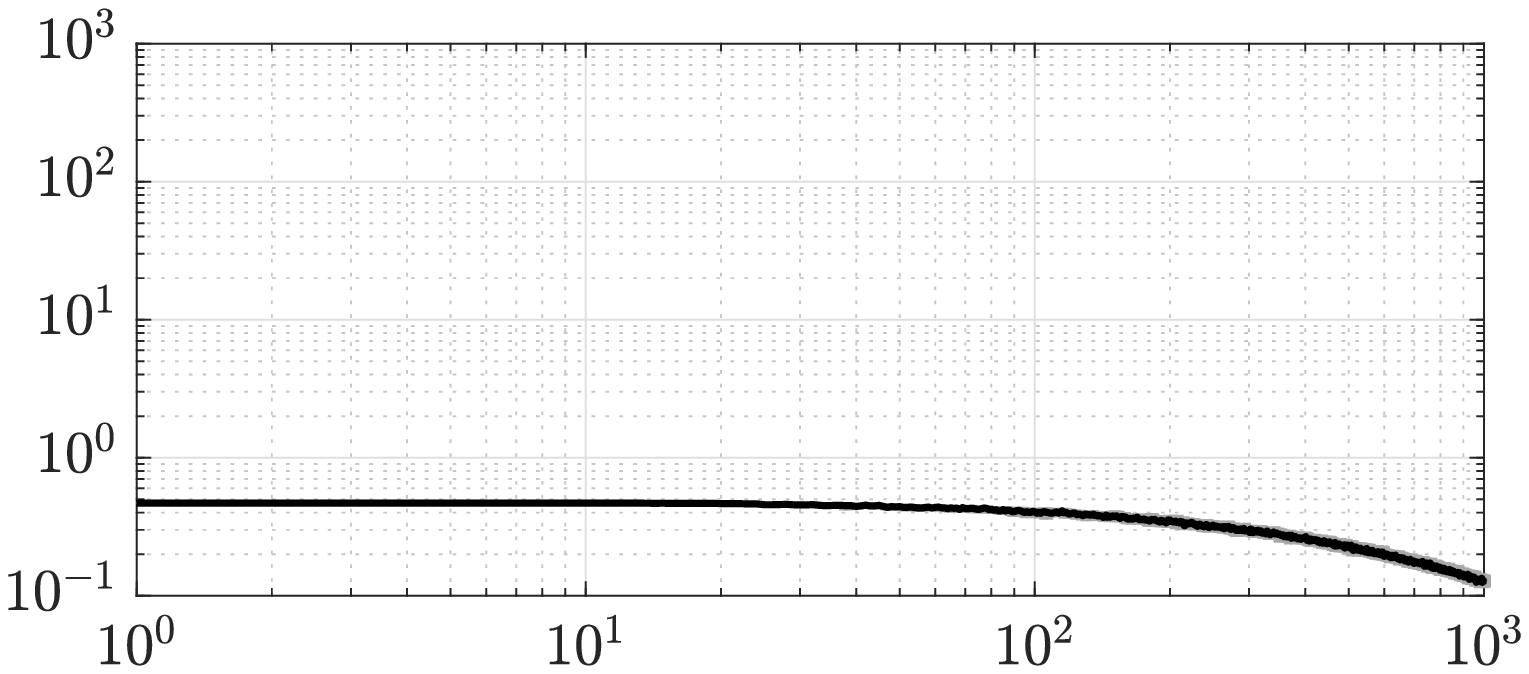}};
		\node[] at (0,-1.9) {\footnotesize $k$};
		\node[rotate=90] at (-4.3,0) {\footnotesize $\psi(\theta)$};
		\node[draw,fill=white,minimum width=3.1cm] at (-1.5,+1.1) {$\epsilon_1=\epsilon_2=\epsilon_3=10$};
		\end{tikzpicture}
	\end{tabular}
	\caption{
		\label{fig:b0}  
		Percentile statistics of relative fitness of 100 runs of Algorithm~\ref{alg:0} for learning length of stay at hospital versus the iteration number $k$ for a learning horizon of $T=1,000$ iterations with three choices of privacy budgets $\epsilon_1=\epsilon_2=\epsilon_3$. The gray area illustrates the range of 25\% to 75\% percentiles for the relative fitness and the black line shows the median of relative fitness. }
\end{figure}

\begin{figure*}
	\centering
	\begin{tikzpicture}
	\node[] at (0,0) {\includegraphics[width=0.98\linewidth]{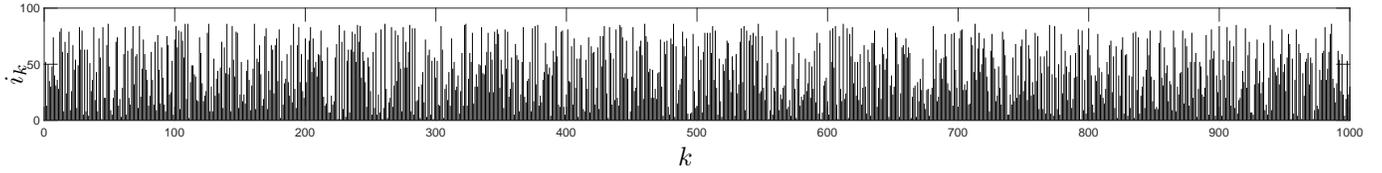}};
	\node[] at (0,-1.1) {$k$};
	\node[rotate=90] at (-8.9,0) {$i_k$};
	\end{tikzpicture}
	\vspace{-.3in}
	\caption{
		\label{fig:b00}
		Example of communication timing for the asynchronous learning in Algorithm~\ref{alg:0} for learning length of stay at hospital, illustrating $i_k$ versus the iteration number $k$.
	}
\end{figure*}

\begin{figure}[t]
	\begin{tikzpicture}
	\node[] at (0,0) {
		\includegraphics[width=1\columnwidth]{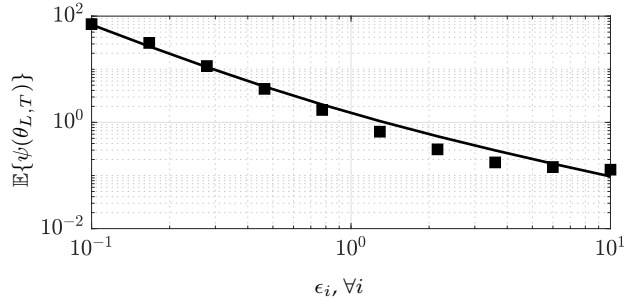}};
	\node[] at (0,-2.1) {\footnotesize $\epsilon_i$, $\forall i$};
	\node[rotate=90] at (-4.2,0) {\footnotesize $\mathbb{E}\{\psi(\theta_{L,T})\}$};
	\end{tikzpicture}
	\vspace{-.2in}
	\caption{\label{fig:b1} Relative fitness of Algorithm~\ref{alg:0} for learning length of stay at hospital after $T=1,000$ iterations versus the privacy budget $\epsilon_i$, $\forall i$. The solid line illustrates the bound in~\eqref{eqn:2_limiting} with $\bar{c}'_1=0.9$ and $\bar{c}'_2=0.6$.}
\end{figure}

Figure~\ref{fig:b2} illustrates the relative fitness of Algorithm~\ref{alg:0} for learning  length of stay at hospital after $T=1,000$ iterations, $\mathbb{E}\{\psi(\theta_{L,T})\}$, for three choices of privacy budgets $\epsilon_i=0.1$ (black line),  $\epsilon_i=1$ (dashed line),  $\epsilon_i=10$ (dash-dotted line). The markers show the relative fitness of the non-private ML model $\theta^*_i$, $\psi(\theta_i^*)$, constructed based on only the private data of the $i$-th data owner versus the size of the data set owned by the $i$-th data owner. For $\epsilon=10$, eight hospitals (i.e., Women And Children's Hospital Of Buffalo, Crouse Hospital, St Peters Hospital, White Plains Hospital Center, Westchester Medical Center, Memorial Hospital for Cancer and Allied Diseases, Long Island Jewish Schneiders Children's Hospital Division, St Francis Hospital) benefit from collaboration. The relative fitness of the non-private ML model $\theta^*_i$ for these eight hospitals are above the dash-dotted line. 

We demonstrate the performance of the iterates of Algorithm~\ref{alg:0}. Figure~\ref{fig:b0} illustrates the percentile statistics of the relative fitness $\psi(\theta_{L,k})$ for 100 runs of Algorithm~\ref{alg:0} versus the iteration number $k$ for the learning horizon $T=1,000$. Figure~\ref{fig:b00} illustrates an example of communication timing for the asynchronous learning in Algorithm~\ref{alg:0}, illustrating $i_k$ versus the iteration number $k$. At each iteration, only one of the 86 data owners communicates with the learner. The gray area in Figure~\ref{fig:b0} shows 25\%--75\% percentiles of the relative fitness and the black solid lines in show the median of relative fitness versus the iteration number. For large privacy budgets, the median decreases across time until the algorithm converges to a neighbourhood of the solution of~\eqref{eqn:ML}. The relative fitness of the trained model also improves as $\epsilon_1=\epsilon_2=\epsilon_3$ increases. 

Figure~\ref{fig:b1} shows the average relative fitness of the trained ML model  using Algorithm~\ref{alg:0} after $T=1,000$ iterations, $\mathbb{E}\{\psi(\theta_{L,T})\}$, versus the privacy budgets  $\epsilon_1=\epsilon_2=\epsilon_3$. The markers (i.e., {\small$\blacksquare$}) show the  experiments. Evidently, the relative fitness rapidly improves as the privacy budgets  $\epsilon_1=\epsilon_2=\epsilon_3$ increase.

\section{Conclusions and Future Research}
\label{sec:conclusions}
In this paper, we developed an asynchronous DP algorithm for training ML models on multiple private datasets. We proved that, by following the asynchronous algorithm, the cost of privacy is inversely proportional to the combined size of the  training datasets squared and the privacy budgets squared. Finally, we validated the theoretical results on experiments on financial data. Future work can focus on multiple directions. An interesting extension is to consider multiple learners training separate  ML models. This would be more similar to the distributed ML on arbitrary connected graphs. This way, we can extend the results to more general communication structures with the learner not necessarily at the center. We can investigate the behaviour of private data owners and learners in a data market. The cost of privacy in this paper can be used as a guide for developing compensation mechanisms for private data owners to increase their privacy budgets. The developed algorithm is particularly of use as the data owners and the learners in the data market can predict the performance of privately-trained ML models during negotiation for setting privacy budgets and compensating data owners. Finally, we can extend the results to adversarial ML with more sophisticated adversaries.

\section*{Acknowledgements}
The work has been funded, in part, by the ``Data Privacy in AI Platforms (DPAIP): Risks Quantification and Defence Apparatus'' project from the Next Generation Technologies Fund by the Defence Science and Technology
(DST) in the Australian Department of Defence and the DataRing project funded by the NSW Cyber Security Network and Singtel Optus pty ltd through the Optus Macquarie University Cyber Security Hub. Parts of this work was conducted while F. Farokhi was affiliated with CSIRO's Data61. He is thankful for their support.

\bibliographystyle{ieeetr}
\bibliography{citation}

\appendices

\section{Proof of Theorem~\ref{tho:1}}
\label{proof:tho:1}
Since there are at most $T$ rounds of communication, the privacy budget in each step should be set as $\epsilon_i/T$ for all $i$. Now, note that  
\begin{align*}
    \|\mathcal{Q}_{i_k}(\mathcal{D}_{i_k};\bar{\theta}_{k})-&\mathcal{Q}_{i_k}(\mathcal{D}'_{i_k};\bar{\theta}_{k})\|_1\\
=&\frac{1}{n_{i_k}}\Bigg\|\sum_{\{x,y\}\in\mathcal{D}_{i_k}} \hspace{-.05in}\nabla_\theta\ell(\mathfrak{M}(x;\theta),y)\\
&\hspace{.1in}-\sum_{\{x,y\}\in\mathcal{D}'_{i_k}}\hspace{-.05in} \nabla_\theta\ell(\mathfrak{M}(x;\theta),y)\Bigg\|_1\\
=&\frac{1}{n_{i_k}}\Big\| \nabla_\theta\ell(\mathfrak{M}(x;\theta),y)|_{\{x,y\}\in\mathcal{D}_{i_k}\setminus\mathcal{D}'_{i_k}}\\
&\hspace{.1in}-\nabla_\theta\ell(\mathfrak{M}(x;\theta),y)
|_{\{x,y\}\in\mathcal{D}'_{i_k}\setminus\mathcal{D}_{i_k}}\Big\|_1\\
=&\frac{2\Xi}{n_{i_k}}.
\end{align*}
Therefore, the scale of the noise must be  $2\Xi T/(n_{i_k}\epsilon_{i_k})$.

\section{Proof of Theorem~\ref{tho:2}}
\label{proof:tho:2}
We start by casting the problem of privacy-aware learning in the framework of asynchronous distributed optimization in~\cite{5454103}. For any $\eta<1/N$, we can define
$
    f_i(\theta)=\eta g(\theta)+\frac{1}{n} \sum_{\{x,y\}\in\mathcal{D}_i} \ell(\mathfrak{M}(x;\theta),y), \forall i\in\mathcal{N},
$
and
$
    f_L(\theta)=(1-\eta N)g(\theta).
$
We can think of $f_i$ as the cost functions of data owners and $f_L$ as the cost function of the learner. By construct, $f_L$ is $\sigma_L$ strongly convex with $\sigma_L=(1-\eta N)\sigma$ and 
$f_i$ is $\sigma_i$ strongly convex with $\sigma_i=\eta\sigma$. Note that
\begin{align*}
    \|\nabla_\theta f_i(\theta)\|_2=
    &\left\|\eta \nabla_\theta g(\theta)+\frac{1}{n}\sum_{\{x,y\}\in\mathcal{D}_i}\nabla_\theta \ell(\mathfrak{M}(x;\theta),y)\right\|\\
    \leq & \eta\Xi_g+\frac{n_i}{n}\Xi\\
    \leq & \Xi_g+\Xi,
\end{align*}
and
$
    \|\nabla_\theta f_L(\theta)\|_2
    =\|(1-\eta N) \nabla_\theta g(\theta)\|
    \leq (1-\eta N)\Xi_g
    \leq  \Xi_g.
$
Therefore, $\|\nabla_\theta f_i(\theta)\|_2\leq C$, $\forall i$, and $\|\nabla_\theta f_L(\theta)\|_2\leq C$ with $C=\Xi_g+\Xi$.

In each iteration, one of the data owners at random is selected and follows the gossip algorithm (see~\cite{5454103}) for exchanging information in learning and updating the decision variables. In this paper, however, we assume that the learner takes care of all the updates and storing the iterates. Therefore, the learner submits a gradient query to the selected data owner and receives a DP response for updating the decision variables. Let $i$ denote the index of the randomly-selected data owner at iteration $k$; note that $i_k$ is used in Algorithm~\ref{alg:0} for denoting the index. We use $\mathcal{G}=(\mathcal{V},\mathcal{E})$ to denote a graph with the vertex set $\mathcal{V}=\{1,\dots,N,N+1\}$, in which node $N+1$ is the learner $L$, and the edge set $\mathcal{E}\subseteq\mathcal{V}\times\mathcal{V}$. By the methodology of~\cite{5454103}, we get
\begin{align*}
    W_k=I-\frac{1}{2}(e_i-e_{N+1})(e_i-e_{N+1})^\top, 
\end{align*}
and $U_k=\{L,i\}$. It is evident that the probability of selecting the learner at each round is equal to one, i.e., $\gamma_L=1$, and the probability of selecting any data owner is $\gamma_i=1/N$ in the notation of~\cite{5454103},. We get
\begin{align*}
    \overline{W}
    &=\mathbb{E}\{W_k\}\\
    &=I-\begin{bmatrix} \displaystyle
    \displaystyle\frac{1}{2N} & 0 & \cdots & 0 & -\displaystyle\frac{1}{2N} \\[.8em]
    0 & \displaystyle\frac{1}{2N} & \cdots & 0 & -\displaystyle\frac{1}{2N} \\[.8em]
    \vdots & \vdots & \ddots & \vdots & \vdots \\[.8em]
    0 & 0 & \cdots & \displaystyle\frac{1}{2N} & \displaystyle\frac{1}{2N} \\[.8em]
    -\displaystyle\frac{1}{2N} & -\displaystyle\frac{1}{2N} & \cdots & -\displaystyle\frac{1}{2N} & \displaystyle\frac{1}{2}
    \end{bmatrix}.
\end{align*}
We meet all the conditions of Assumption 2 in~\cite{5454103}. Furthermore, using Theorem 1 in~\cite{5454103}, we can see that 
\begin{align*}
    \lambda=\left\|W_k-\frac{1}{N+1}\mathds{1}\mathds{1}^\top W_k\right\|_2^2<1.
\end{align*}
The updates in~(2) in~\cite{5454103} can be rewritten as 
\begin{align*}
    \bar{\theta}_{k}=\frac{1}{2}\theta_{L,k-1}+\frac{1}{2}\theta_{i,k-1},
\end{align*}
with the notation substitution of $\bar{\theta}_{k}$ instead of $v_{i,k}=v_{L,k}$, $\theta_{i,k}$ instead of $x_{i,k}$, and $\theta_{L,k}$ instead of $x_{L,k}$. The updates in (3) in~\cite{5454103} can also be rewritten as 
\begin{align*}
    \theta_{i,k}
    =&\Pi_\Theta\left[\bar{\theta}_{k}-\alpha_i\eta\nabla_\theta g(v_{k})+\alpha_i\frac{n_i}{n}\overline{\mathcal{Q}}_i(\bar{\theta}_{k};k)\right]\\
    =&\Pi_\Theta\left[\bar{\theta}_{k}-\alpha_i\left(\eta\nabla_\theta g(v_{k})+\frac{n_i}{n}\left(\mathcal{Q}_i(\bar{\theta}_{k};k)+w_{i}(k)\right)\right) \right]\\
    =&\Pi_\Theta\left[\bar{\theta}_{k}-\alpha_i(\nabla_\theta f_i(\bar{\theta}_{k})+\bar{w}_{i}(k)) \right],
\end{align*}
with $\bar{w}_{i}(k)=w_{i}(k)n_i/n$ and 
\begin{align*}
    \theta_{L,k}=&\Pi_\Theta\left[\bar{\theta}_{k}-\alpha_L\nabla f_L(\bar{\theta}_{k}) \right]\\
    =&\Pi_\Theta\left[\bar{\theta}_{k}-(1-\eta N)\alpha_L\nabla_\theta g(\bar{\theta}_{k}) \right],
\end{align*}
where $w_{i}(k)$ is the additive i.i.d. DP noise and
\begin{align*}
    \mathcal{Q}_i(\bar{\theta}_{k};k)=\frac{1}{n_i} \sum_{\{x,y\}\in\mathcal{D}_i} \nabla_\theta\ell(\mathfrak{M}(x;\bar{\theta}_{k}),y).
\end{align*}
Note that, here, we are using $i$ instead of $i_k$ to reduce the complexity of the notation and for conforming to the notation of~\cite{5454103}. We have
\begin{align*}
    \mathbb{E}\{\bar{w}_{i}(k)|\mathcal{F}_k\}&=0,\\
    \mathbb{E}\{\|\bar{w}_{i}(k)\|_2^2|\mathcal{F}_k\}&\leq \nu_i^2,
\end{align*}
where $\mathcal{F}_k$ is the filtration generated by the entire history of Algorithm~\ref{alg:0} up to iteration $k$. Using Theorem~\ref{tho:1}, we can see that
\begin{align*}
    \nu_i=\frac{2\sqrt{2}\Xi T}{n\epsilon_i}.
\end{align*}
Extending Lemma 3 in~\cite{5454103} results in
\begin{align*}
    \mathbb{E}\{\|\nabla_\theta f_i(\bar{\theta}_k)+w_{i}(k)\|_2^2|\mathcal{F}_{k-1},W_k\}
    &\leq C^2+\nu_i^2\leq (C+\nu_i)^2,\\
    \mathbb{E}\{\|\nabla_\theta f_L(\bar{\theta}_k)\|_2^2|\mathcal{F}_{k-1},W_k\}&\leq C^2.
\end{align*}
Therefore, we can upgrade the right-hand side of (22) in~\cite{5454103} to 
\begin{align*}
    \mathbb{E}\{\alpha_i^2(C+\nu_i)^2\}+\alpha_L^2C^2
    =\alpha_L^2C^2+\frac{1}{N}\sum_{i\in\mathcal{N}}\alpha_i^2(C+\nu_i)^2
\end{align*}
Note that, in the case of this paper, the summation only contains two terms because, in each iteration, only the learner and another data owner update their decision variables. This implies that, in Proposition 1 in~\cite{5454103}, $\varepsilon_{\mathrm{net}}$ must be updated to
\begin{align*}
    \varepsilon_{\mathrm{net}}=\frac{C\sqrt{N+1}}{1-\sqrt{\lambda}}\sqrt{\alpha_L^2+\frac{1}{N}\sum_{i\in\mathcal{N}}\alpha_i^2\left(1+\frac{\nu_i}{C}\right)^2}.
\end{align*}
With the same line of reasoning, we can improve the bound in Proposition 2 in~\cite{5454103} to get
\begin{align}
    &\limsup_{k\rightarrow \infty} \left[ \mathbb{E}\{\|\theta_{L,k}-\theta^*\|_2^2\}+\sum_{i\in\mathcal{N}}\mathbb{E}\{\|\theta_{i,k}-\theta^*\|_2^2\}\right]\nonumber\\
    &\hspace{1.55in}\leq \frac{\varepsilon+2\alpha_{\max}C\varepsilon_{\mathrm{net}}}{1-q},
\end{align}
where
\begin{align} 
    \varepsilon=&2(N+1)(1-\gamma_{\min})\delta_{\alpha,\sigma}\diam(\Theta)^2\nonumber\\
    &2(N+1)\delta_{ \alpha,\gamma}C\diam(\Theta)\nonumber\\
    &+C^2\bigg(\alpha_L^2+\frac{1}{N}\sum_{i\in\mathcal{N}}\alpha_i^2\left(1+\frac{\nu_i}{C}\right)^2\bigg),\label{eqn:varepsilon}
\end{align}
and
\begin{align*}
    \alpha_{\max}=&\max_i \alpha_i,\\
    \gamma_{\min}=&1/N,\\
    q=&1-2\gamma_{\min}\min\left\{\alpha_L\sigma_L,\min_{i\in\mathcal{N}}\alpha_i\sigma_i\right\},\\
    \delta_{\alpha,\sigma}=&\max\hspace{-.03in}\left\{\hspace{-.03in}\alpha_L\sigma_L,\max_{i\in\mathcal{N}}\alpha_i\sigma_i\hspace{-.03in}\right\}\hspace{-.03in}-\hspace{-.03in}\min\hspace{-.03in}\left\{\hspace{-.03in}\alpha_L\sigma_L,\min_{i\in\mathcal{N}}\alpha_i\sigma_i\hspace{-.03in}\right\}\hspace{-.03in},\\
    \delta_{\alpha,\gamma}=&\max\hspace{-.03in}\left\{\hspace{-.03in}\alpha_L\gamma_L,\max_{i\in\mathcal{N}}\alpha_i\gamma_i\hspace{-.03in}\right\}\hspace{-.03in}-\hspace{-.03in}\min\hspace{-.03in}\left\{\hspace{-.03in}\alpha_L\gamma_L,\min_{i\in\mathcal{N}}\alpha_i\gamma_i\hspace{-.03in}\right\}\hspace{-.03in}.
\end{align*}
Therefore, for any $\varsigma>0$, there exists large enough $T\in\mathbb{N}$ such that 
\begin{align}
    &\left[ \mathbb{E}\{\|\theta_{L,T}-\theta^*\|_2^2\}+\sum_{i\in\mathcal{N}}\mathbb{E}\{\|\theta_{i,T}-\theta^*\|_2^2\}\right]\nonumber\\
    &\hspace{1.35in}\leq \varsigma+\frac{\varepsilon+2\alpha_{\max}C\varepsilon_{\mathrm{net}}}{1-q},
    \label{eqn:bound}
\end{align}
Selecting $\eta=1/(2N)$ and $\alpha_L=\alpha_i/N=\alpha/\sigma$ for some constant $\alpha\in(0,1)$, we get $\delta_{\alpha,\sigma}=\delta_{\alpha,\gamma}=0$. 
Therefore, we can simplify~\eqref{eqn:varepsilon} to get 
\begin{align}\label{eqn:varepsilon1}
    \varepsilon=&\frac{\alpha^2C^2}{\sigma^2}\bigg(1+N\sum_{i\in\mathcal{N}}\left(1+\frac{\nu_i}{C}\right)^2\bigg).
\end{align}
We will also get
\begin{align}\label{eqn:varepsilonnet}
    2\alpha_{\max}C\varepsilon_{\mathrm{net}}=&\frac{2N\alpha^2 C^2\sqrt{N+1}}{\sigma^2(1-\sqrt{\lambda})}\nonumber\\
    &\times \sqrt{1+N\sum_{i\in\mathcal{N}}\left(1+\frac{\nu_i}{C}\right)^2}.
\end{align}
Furthermore, 
\begin{align}\label{eqn:q}
    1-q=2\gamma_{\min}\min\left\{\alpha_L\sigma_L,\min_{i\in\mathcal{N}}\alpha_i\sigma_i\right\}=\frac{\alpha}{N}.
\end{align}
Combining~\eqref{eqn:bound} with~\eqref{eqn:varepsilon1}--\eqref{eqn:q}, we get
\begin{align*}
    \mathbb{E}&\{\|\theta_{L,T}-\theta^*\|_2^2\}\\
    \leq & \left[ \mathbb{E}\{\|\theta_{L,T}-\theta^*\|_2^2\}+\sum_{i\in\mathcal{N}}\mathbb{E}\{\|\theta_{i,T}-\theta^*\|_2^2\}\right]\\
    \leq &\frac{N\alpha C^2}{\sigma^2}\left(1+N\sum_{i\in\mathcal{N}}\left(1+\frac{2\sqrt{2}\Xi T}{n\epsilon_i(\Xi_g+\Xi)}\right)^2\right)\\
    &+\frac{2N^2\alpha C^2\sqrt{N+1}}{\sigma^2(1-\sqrt{\lambda})} \sqrt{1\hspace{-.03in}+\hspace{-.03in}N\hspace{-.03in}\sum_{i\in\mathcal{N}}\hspace{-.03in}\left(\hspace{-.03in}1\hspace{-.03in}+\hspace{-.03in}\frac{2\sqrt{2}\Xi T}{n\epsilon_i(\Xi_g+\Xi)}\right)^{\hspace{-.05in}2}}.
\end{align*}
Define 
$
    c_1={NC^2}/{\sigma^2},$ $c_2={2N^2 C^2\sqrt{N+1}}/({\sigma^2(1-\sqrt{\lambda})}).
$
We have
\begin{align*}
\mathbb{E}\{\|\theta_{L,T}&-\theta^*\|_2^2\}\\
    \leq &c_1\alpha\hspace{-.03in}\left(\hspace{-.03in}1\hspace{-.03in}+\hspace{-.03in}N\sum_{i\in\mathcal{N}}\left(1+\frac{2\sqrt{2}\Xi T}{n\epsilon_i(\Xi_g+\Xi)}\right)^{\hspace{-.03in}2}\right)\\
    &+c_2\alpha\sqrt{1\hspace{-.03in}+\hspace{-.03in}N\sum_{i\in\mathcal{N}}\left(\hspace{-.03in}1+\frac{2\sqrt{2}\Xi T}{n\epsilon_i(\Xi_g+\Xi)}\right)^{\hspace{-.03in}2}}.
\end{align*}
Selecting $\alpha=\rho/T^2$ and noting that $\Xi\leq \Xi_g+\Xi$, the upper bound can be further simplified~\eqref{eqn:bound:1}. 
Following the same modifications in the proof of Proposition~3 in~\cite{5454103} results in~\eqref{eqn:bound:2}. 

\end{document}